# ARTIFICIAL INTELLIGENCE FOR SUSTAINABLE WINE INDUSTRY: AI-DRIVEN MANAGEMENT IN VITICULTURE, WINE PRODUCTION AND ENOTOURISM

Marta Sidorkiewicz[a], Karolina Królikowska[b], Berenika Dyczek[c], Edyta Pijet-Migoń[d], Anna Dubel[e]

[a] University of Szczecin, Institute of Management, Szczecin, Poland
[b] Wroclaw University of Environmental and Life Sciences, Institute of Spatial Management, Wroclaw, Poland
[c] University of Wroclaw, Institute of Sociology, Wroclaw, Poland
[d] WSB Merito University in Wroclaw, Faculty of Finance and Management, Wroclaw, Poland
[e] AGH University of Krakow, Faculty of Management, Poland

## ABSTRACT

**Purpose:** *This study examines the role of Artificial Intelligence (AI) in enhancing sustainability and efficiency within the wine industry. It focuses on AI-driven intelligent management in viticulture, wine production, and enotourism.*

**Need for the Study:** *As the wine industry faces environmental and economic challenges, AI offers innovative solutions to optimize resource use, reduce environmental impact, and improve customer engagement. Understanding AI's potential in sustainable winemaking is crucial for fostering responsible and efficient industry practices.*

**Methodology:** *The research is based on a questionnaire survey conducted among Polish winemakers, combined with a comprehensive analysis of AI methods applicable to viticulture, production, and tourism. Key AI technologies, including predictive analytics, machine learning, and computer vision, are explored.*

**Findings:** *AI enhances vineyard monitoring, optimizes irrigation, and streamlines production processes, contributing to sustainable resource management. In enotourism, AI-powered chatbots, recommendation systems, and virtual tastings personalize consumer experiences. The study underscores AI's impact on economic, environmental, and social sustainability, supporting local wine enterprises and cultural heritage.*

**Practical Implications:** *AI in winemaking and enotourism can lead to more efficient, sustainable operations that benefit producers and consumers. AI-driven solutions promote responsible tourism, enhance wine tourism experiences, and ensure the industry's long-term viability.*

## 1. INTRODUCTION

Sustainability in the wine industry encompasses environmental stewardship, economic viability, and social responsibility. Sustainable viticulture aims to minimize environmental impacts while maintaining product quality. Environmental issues have become very important among winemakers. They are strongly linked to corporate image (Santini et al., 2013), especially when winery customers push for environmentally friendly practices (De Castro et al., 2024). Additionally, the adoption of organic and biodynamic farming methods has increased in recent years (Baiano, 2021), contributing to the overall sustainability of vineyards. In wine production, sustainability efforts focus on reducing resource consumption and waste or implementing energy-efficient technologies. This approach contributes to reducing carbon emissions, which the wine industry is increasingly compelled to do by market and regulatory factors that force it to assess, reduce and communicate its carbon footprint (da Silva & da Silva, 2022). In addition to viticulture and wine production, an important part of the wine industry is enotourism. Many authors emphasize that wine tourism may become an additional source of income and contribute to the preservation of cultural heritage and cultural wine landscapes in regions facing various economic and demographic problems (Lourenço-Gomes, Pinto & Rebelo, 2015; Montella, 2017; Senkiv et al., 2022). Considerably fewer studies address wine tourism in a complex way, paying equal attention to the three pillars of sustainable development – economic, environmental and social impact. Among them is undoubtedly the report „Sustainable Wine Tourism: A Global Survey" (Szolnoki et al., 2022). After a survey involving indices of sustainable development, these authors assert that wine tourism could be the key element for the sustainable development of wine regions worldwide. Many studies highlight that sustainability in wine tourism is increasingly linked to competitive advantages, influencing consumer preferences and industry practices (Martínez-Falcó et al., 2024). Despite advancements, challenges remain in fully integrating sustainability into the wine industry. Future research should focus on developing innovative solutions to these challenges, including applying artificial intelligence solutions (Newlands, 2021).

The primary objective of this study is twofold: (a) to present the findings of a survey conducted among Polish winemakers regarding their approach to sustainability, and (b) to identify potential applications of artificial intelligence (AI) that can enhance sustainability in viticulture, wine production, and enotourism. The study is grounded in the three pillars of sustainability - environmental, social, and economic - and seeks to comprehensively analyze current attitudes in the Polish wine industry and their potential for AI-driven innovation.

The research was designed as a full-population study rather than a sample-based survey, as the questionnaire was distributed to all officially registered vineyards in Poland at the time of the study. This methodological approach allows for capturing the Polish wine industry in the initial stage. In the environmental dimension, the study examines pro-environmental viticulture methods, sustainable wine production practices, and the availability of organic and natural wines. The social dimension focuses on the role of wineries in shaping the cultural landscape, their architectural impact, and their contributions to local communities. The economic dimension explores how viticulture, wine production, and enotourism contribute to regional economic development.

The analysis of collected data suggests that Poland's relatively small-scale production and limited vineyard acreage—characteristic of an early-stage winemaking industry—create favorable conditions for sustainable development. However, potential expansion and a shift toward mass production may challenge the sector's sustainability. Consequently, the study discusses how AI-driven solutions can help maintain environmentally and socially responsible winemaking practices.

## 2. LITERATURE REVIEW

Vineyards and viticulture are closely linked to a territory and its natural and cultural heritage. In Poland, it is assumed that the development of viticulture, as a cultural phenomenon and an economic enterprise, can be divided into three distinct phases: from the adoption of Christianity (with the culmination of development at the turn of the thirteenth and fourteenth centuries) to the decline of the First Polish Republic, the 20th century (especially the 1930s and 1950s) and the Third Polish Republic and the present day (Jęczmyk, 2019).

Grape growing was widespread in Poland centuries ago. The oldest evidence of vines found on the hillsides of the Wawel Royal Castle in Kraków dates from the turn of the 9th and 10th centuries, but proper development began in the 12th century (Dylik-Ostrowska et al., 2021; Fundacja Galicja Vitis, 2017; Kapczyński, 2021). Grapes came to Poland with Christianity, and the first winemakers were the Benedictine and Cistercian monks, who cultivated these plants on monastery farms and used the wine for liturgical purposes (Wawro, 2011). At first, viticulture in Poland was not easy, as it required previously unknown techniques, and the work could only be done by adequately trained people. It is believed that this is why vine-growing first took place on land belonging to churches, monasteries, or princes. Only later did vineyards appear near hospitals, as wine was used to cure various diseases (Myśliwiec, 2013). In the 13th century, winemaking was of little economic importance in Poland and wine was the luxury drink of the wealthy. It was not until the 14th century that a revolutionary change took place and this was when this activity became most widespread in Poland. At this time, wine-making was also taken up by the bourgeoisie, who saw that it was a profitable profession (Poczta &

Zagrocka, 2016). The crisis of Polish winemaking began in the 16th century, when wine began to be imported from countries where viticulture was much easier (mainly from Hungary, France and Italy), and Polish wine slowly gave way to cheaper foreign wine. However, this was not the only factor that caused a decline in interest in setting up one's vineyards. Unfavorable climatic conditions were also a contributing factor (including the so-called Little Ice Age), effectively destroying the crop, making cultivation more expensive and less profitable. However, the most significant damage to Polish vineyards was caused by wars in the 17th and 20th centuries (Dylik-Ostrowska et al., 2021; Myśliwiec, 2013; Olewnicki, 2018). The phylloxera plague decimated vineyards across Europe, another factor stopping Poland's winemaking development.

World War II was a time of devastation for vineyards and the moving borders on the map of Europe. The post-war authorities initially planned to rebuild the wine infrastructure, but the freezing winter of 1956/1957 made the planned winemaking ventures seem doomed to failure. More important were the large, socialist state farms, the collectivization of agriculture through its control by the state and the forced amalgamation of individual farms into huge cooperatives, and the distraction from the mistakes made with this. It was not until the collapse of communism that economic and cultural changes revived history and made changes possible (Kapczyński, 2021).

The current history of wine making in Poland began in the second half of the 20th century, in the 1980s. (Dylik-Ostrowska et al., 2021; Wawro, 2011). The growing interest in food, cooking, culinary programs and the fashion for preparing meals together has also led to an increased interest in beverages, including wine, which is often not only a component of a dish but also an essential accompaniment to a meal (Mazurkiewicz-Pizło, 2010). The warming of the climate and the successive introduction of new vine varieties, resistant or less sensitive to disease and above all resistant to frost (known as hybrids), make it possible to establish vineyards in areas with a more difficult climate (Olewnicki, 2018). Not insignificant is the legislation, which is changing in favor of winemakers, concerning the conduct of broadly defined winemaking activities. The development of winemaking in Poland is the most dynamic in Poland's history, and there are more and more vineyards every year. New vineyards are being established not only in regions long associated with wine, but all over the country, an example of which is the north-western region of Zachodniopomorskie.

The area of cultivated vineyards in Poland varies from a dozen or so acres to several dozen hectares. The largest vineyard in Poland is the Turnau Vineyard, which covers 41 hectares. (Winnica Turnau, 2024). Polish vineyards also have a diverse character - they are family-owned, non-commercial vineyards, and organized enterprises conducting business activities.

A relatively thorough basis for understanding domestic wine production is the lists of the National Agricultural Support Centre (Krajowy Ośrodek Wsparcia Rolnictwa KOWR, Wykazy i Rejestry, 2023). In 2023, there were 530 registered entrepreneurs authorized to engage in the business of making or bottling wine products. However, this number does not reflect the phenomenon's scale under analysis. According to popular internet portals run by winemakers, in March 2023, over 580 vineyards were operating in Poland. The relatively large number discrepancy has to do with the fact that not all winemakers register their winemaking activities, some vineyards are tiny and produce wine only for their own needs, and others are very young plantings where the vines do not yet produce adequate fruit. Three regions can, however, be recognized as Poland's wine-growing centers in terms of the number of vineyards and acreage planted: Małopolskie, Lubuskie and Dolnośląskie, which have the highest number of wine producers reported to the KOWR for the 2022/2023 wine year - 77, 55 and 51, with 103, 115 and 78 hectares respectively.

Observing the Polish wine market, one can conclude that it is developing quickly. Such a situation is probably also correlated with the fact that the preferences of Polish consumers regarding the consumption of alcoholic beverages are changing. The percentage of persons most willingly choosing wine has been increasing since 2007 (Polacy piją mniej piwa, więcej wina i alkoholi wysokogatunkowych, 2019). From November 2019 to October 2020, wine retail sales increased year-on-year by 9.9% in value terms and by 6.5% in volume terms by number of bottles and by 6.1% in liters (W Polsce, krainie wódki i piwa, pije się coraz więcej wina, 2020). The consequence of such a state of affairs seems to be the further development of winemaking in Poland.

## 3. METHODS

The questionnaire survey was performed electronically, using the Microsoft 365 platform. 75 vineyards participated in the survey. The questionnaire was accessible from May 2022 until February 2023. The invitation to participate in the survey was sent, along with the link to the online questionnaire, to all vineyards registered in the official Register of Vineyards managed by the National Support Centre for Agriculture. It was either involved in wine sales in the economic year in agriculture 2021/2022 or intended to do so. The register included 376 vineyards, yet it was subject to verification using other sources of information, wine-related Internet portals and the authors' previous research. In effect, considering repetitions (several plots in different localities belonging to one vineyard) and vineyards that terminated their activity, the number of active vineyards involved in vine cultivation and wine selling was reduced to 308. Requests to fill out the questionnaire were sent to all these vineyards. Accordingly, we undertook a comprehensive survey of all active vineyards in the country, which provides a substantial sample of the enotourism business at this nascent stage of development, despite the 24% response rate. While a higher response rate is typically sought in quantitative studies, achieving a sample representing at least 20% of the population is sufficient to draw meaningful insights within sociological research.

Contact data were retrieved from the Enotourist Guide "Polish Wine" (Dylik-Ostrowska et al., 2021), wine-related Internet portals, official websites of vineyards, and social media. Since the questionnaire return rate was relatively small, reminder letters were sent twice in September and October. In addition, direct phone calls and informal meetings with wine owners were used to encourage participation.

Regarding qualitative data analysis, we employed a deductive approach to coding, with guidance from the principles of qualitative research methodologies set forth by Creswell (2013) and Patton (2002). The responses to the open-ended questions were subjected to systematic coding according to the themes that emerged from the survey questions. The initial codes were defined by the structure of the questions, and were subsequently refined based on the responses received. This approach enabled the data to be categorised effectively. A numerical code was assigned to each theme and subsequently linked to the relevant fragments of respondents' answers. The deductive coding approach provided a structured method for analysing qualitative data, allowing for a systematic comparison of themes across responses and facilitating a comprehensive understanding of enotourism dynamics in Polish vineyards.

A comprehensive overview of AI solutions was then carried out, systematically assigning them to the various aspects of sustainability identified by winemakers, including environmental impact reduction, economic efficiency enhancement, and social responsibility improvements.

## 4. RESULTS

### 4.1 Environmental aspects of wine production in Poland

Regarding environmental aspects of wine making and viticulture, respondents were asked if they consider environmental issues in wine production. Of the 75 respondents in our research, 58 (77%) indicated that they do, 17 said it is hard to say, and no one chose a negative answer.

In the follow-up question related to providing examples of such actions, the following aspects were mentioned:

A. Absence/reduction of crop protection products (16 indications)
B. Renewable energy sources (photovoltaics, heat pumps, windmills) (4 indications)
C. Economic management of materials like glass, cardboard, plastic (8 indications)
D. Utilization of post-production organic waste (yeast, seeds, peels, pomace) (19 indications)
E. Organic and/or moderately applied fertilizers (4 indications)
F. Proper water management (9 indications)
G. Organic and/or vegan products in winemaking (e.g., yeast) or no/limited use of chemicals in oenological processes like low-intervention wines (16 indications)
H. Harvesting and processing manually (3 indications)
I.. Use of natural cork (3 indications)
J. Other (13 indications)

In the category "Other," a few general answers could not be coded (like "we care about the environment. However, some stated that the law undertakes activities, e.g., using substances authorized for use. This can hardly be considered a specific pro-environmental measure because it simply means compliance with legal requirements. Table 1 shows the proposed scope of AI applications and possible methods linked to the winemakers' area of interest related to environmental issues.

**Table 1**. Possible scope and methods of AI application in the environmental aspects of the wine industry.

| Winemakers' areas of interest | Scope of AI applications | AI Methods & Tools |
|---|---|---|
| Absence/reduction of crop protection products<br><br>Organic and/or moderately applied fertilizers<br><br>Harvesting and processing manually | Vineyard Cultivation Management | • AI-Driven Drone Monitoring of Vineyards:<br>• *Multispectral Imaging from drones – analyzing vineyard health using images captured in different light spectra.*<br>• *Image Segmentation with CN – detecting vine diseases through neural networks for image processing.*<br>• Intelligent Irrigation Systems:<br>• *IoT-based Water Management – AI processes data from soil moisture sensors to adjust irrigation.*<br>• *Weather Prediction AI – Weather forecasting helps optimize irrigation strategies.*<br>• Predictive Systems for Yield Forecasting<br>• Yield Prediction with ML – analyzing historical data to predict future harvests.<br>• Satellite Data Fusion – AI integrates data from various sources to assess plant health more accurately. |
| Renewable energy sources<br><br>Economic management of materials like glass, cardboard, plastic<br><br>Utilization of post-production organic waste<br><br>Organic and/or vegan products in winemaking<br><br>Harvesting and processing manually | AI am in Wine Production Optimization | • Automation of the Fermentation Process:<br>• *AI-driven Process Control – AI monitors and adjusts real-time fermentation conditions.*<br>• *Time Series Analysis for Fermentation Monitoring – AI detects anomalies in the fermentation process.*<br>• Wine Quality Control through Chemical Analysis:<br>• *AI-enhanced Spectroscopy – chemical analysis of wine composition to detect contaminants.*<br>• *Neural Networks for Wine Classification – classifying wine quality based on chemical patterns.*<br>• Minimization of Production Waste:<br>• *AI-driven Waste Optimization – AI analyzes ways to reuse waste materials.*<br>• *Predictive Maintenance & Waste Reduction – AI forecasts moments of increased waste in the production process.* |
| Renewable energy sources<br><br>Economic management of materials like glass, cardboard, plastic<br><br>Proper water management | Carbon and Water Footprint of Production | • AI-based Energy Monitoring – AI tracks energy consumption and recommends ways to reduce $CO_2$ emissions. |
| Harvesting and processing manually | | |
| Absence/reduction of crop protection products | Biodiversity Management | • AI-driven Biodiversity Monitoring – AI monitors the impact of cultivation on ecosystems. |
| Organic and/or moderately applied fertilizers | | |
| Harvesting and processing manually | | |
| Site choice | Sustainable Spatial Planning | • AI-based GIS Analysis – AI analyzes topography and historical data to identify optimal locations for new vineyards. |

*Source*: own elaboration based on: Parvini, 2025; How AI can boost Europe's wine industry, 2025; Applications of AI in Agriculture to Reduce Carbon Emissions, 2024; Pizzuto, 2023; Jurišić, Stanisavljević & Plaščak, 2010.

*4.2 Economic aspects of wine production in Poland*

Economic aspects of wine tourism relate to supporting local economies, quality of life and employment. Of the 75 respondents in our research, 36 respondents (48%) indicated that they very often deliberately take actions to enhance the local economy and economic well-being of the local citizens (e.g., by employing locals, using local services, buying local products). At the same time, 24 respondents (32%) indicated that they sometimes take such actions, and 12 (16%) indicated that they do them from time to time. Only 3 survey participants (4%) said they have never taken such actions. The responses altogether presented very positive and proactive attitudes towards local economy enhancement. The vast majority (96%) indicated that they contribute to improving local development and only 4% denied taking any action on the matter.

In the follow-up question related to providing examples of such actions, the following aspects were mentioned:
   A. Employment provision for the local people, employing locals (23 indications)
   B. Purchasing from local stores or entrepreneurs, buying local products (38 indications)
   C. Using local services (23 indications)
   D. Promoting local products and services (7 indications)
   E. Providing our products for local shops (2 indications)

The respondents took the suggestions on the possible influence on the local economy from the former question. In the answers, local cooperation and support were often manifested by buying local products and services, followed by employing local people. Moreover, promoting local products and services and providing our products for local shops were also indicated. Table 2 shows the proposed scope of AI applications and possible methods linked to the winemakers' area of interest related to economic issues.

**Table 2**. AI Applications in Wine Business Management Focused on Local Economic Development.

| Winemakers' areas of interest | Scope of AI applications | AI Methods & Tools |
|---|---|---|
| Employment provision for local people<br><br>Using local services | AI-Supported Workforce Planning | • AI-Powered Hiring Platforms: AI-based systems optimize recruitment by matching local job seekers with wineries based on skills and availability.<br>• Chatbot-Based Training Systems: AI-driven chatbots provide vineyard employees with localized training and skill development programs. |
| Purchasing from local stores or entrepreneurs | AI-Powered Local Supply Chain Optimization | • AI-Driven Procurement Matching: AI connects wineries with local suppliers by analyzing pricing, availability, and sustainability factors.<br>• Geospatial AI for Sourcing Local Products: AI maps regional suppliers and recommends optimal local sources for vineyard needs. |
| | | • Examples:<br>• *Llamasoft Supply Chain Guru – AI for supply chain modeling and optimization.*<br>• *ArcGIS AI – AI-powered geospatial analysis for supplier location optimization.*<br>• *Google Earth Engine AI – AI-based geospatial analysis tool for supply chain planning.*<br>• *SAP Ariba AI – AI-driven procurement optimization for local sourcing.*<br>• *Coupa AI Procurement – AI-powered procurement management and automation.* |
| Providing products to local shops | AI-Powered Smart Distribution | • AI-Assisted Route Optimization: AI minimizes transportation costs and carbon footprint by optimizing delivery routes for local wine distribution. Examples:<br>• *ORTEC Route Optimization: This tool utilizes AI to optimize delivery routes, aiming to reduce fuel consumption and improve delivery efficiency*<br>• *UPS ORION: UPS's On-Road Integrated Optimization and Navigation (ORION) system uses AI to optimize delivery routes, significantly reducing fuel consumption and improving delivery times.* |
| Promoting local products and services | AI-Enhanced Community-Based Commerce | • AI for Local Business Visibility: AI enhances winery partnerships with local businesses by identifying collaboration opportunities through data analysis. Examples:<br>• *IBM Watson Natural Language Processing (NLP): This tool analyzes customer opinions by interpreting text data, providing insights into sentiment and intent.*<br>• *Hootsuite AI: This platform offers AI-driven social media trend analysis, helping businesses monitor and respond to emerging trends effectively.*<br>• *Sprinklr AI: Utilizing artificial intelligence, Sprinklr provides comprehensive social media analytics to identify patterns and trends across various platforms.*<br>• *Sentiment Analysis for Local Branding: AI evaluates online customer reviews to tailor marketing strategies supporting local brands.* |
| Fair Trade and Ethical Consumption Awareness | AI-Powered Consumer Awareness Tools | • Fair Trade Chatbots: AI-powered assistants educate consumers about local and ethical wine production through personalized recommendations. Example:<br>• *A language model developed by OpenAI that can be implemented as a chatbot to educate consumers about ethical practices in wine production. It can answer questions regarding the origin of wine, Fair Trade practices, and sustainable agriculture.* |

*Source*: own elaboration based on: Choi & Lim, 2019; Sanders & Wood, 2020; Fosso Wamba et al, 2021; Leonard, 2021; Fontanella, 2024; HireVue, 2025; Intercom, 2025; Coupa, 2025; Esri, 2025; Google Earth Engine, 2025; Sap, 2025; Ortec, 2025; Ibm, 2025; Hootsuite, 2025; Sprinkl, 2025.

*4.3 Social aspects of wine production in Poland*

Social relations, including those with residents of the area surrounding the vineyard are significant from the perspective of wine industry sustainability. Our survey research indicates that vineyard owners rate their relations with these residents highly: 66 out of 75 responses were either 'good' or 'rather good,' with no responses indicating 'rather bad' or 'very bad.'

It is also crucial to examine the specific nature of the cooperation between winemakers and residents and identify potential conflicts. Such conflicts are particularly concerning in sustainable tourism and community relations. To this end, we analyzed answers to open-ended questions. The question on positive cooperation with residents surrounding the vineyard was explored through an open-ended question: "Can you provide examples of good cooperation with residents in the vineyard area?" This question received responses from 63 respondents.

The responses were categorized into the following content areas:

A. Community Involvement and Support (e.g. mutual equipment support, shared assistance, residents joining in grape harvesting, local help with vineyard tasks, and mutual lending of agricultural equipment (41 indications)

B. Brand and Business Development (e.g. job creation and financial support for those in need) (8 indications).

C. Collaboration with Local Businesses and Organizations (e.g. partnerships with local authorities, cultural organizations, gastronomy, hoteliers, foundations, women's associations, transport companies, and tour guides) (8 indications).

D. Social and Cultural Participation (e.g. engagement in local cultural activities, organizing winery events, and numerous community initiatives (4 indications).

E. No Cooperation: 2 responses indicated no cooperation.

The proposed scope of AI applications and possible methods linked to the winemakers' area of interest related to social issues are shown in Table 3.

**Table 3.** AI Applications in Community Engagement, Business Development, and Cultural Participation in the Wine Industry.

| Winemakers' areas of Interest | Scope of AI applications | AI Methods & Tools |
|---|---|---|
| Community Involvement & Support (e.g., mutual equipment support, shared assistance, grape harvesting) | AI-Powered Resource Sharing Platforms | • Blockchain-Enabled Equipment Sharing: AI-based platforms utilizing blockchain technology for secure and transparent sharing of agricultural machinery.<br>• AI-Driven Task Coordination: AI schedules and optimizes shared labor and resource allocation.<br>• AI-Powered Smart Volunteering Platforms: AI facilitates task-matching for vineyard work, encouraging local participation.<br>• Examples:<br>• *AgriChain (blockchain for supply chain management)*.<br>• *Taranis (AI for precision agriculture & labor coordination)*.<br>• *Grapeworks (AI vineyard management)*. |
| Brand & Business Development (e.g., job creation, crowdfunding) | AI-Enhanced Marketing & Customer Engagement | • AI-Driven Customer Insights: AI analyzes market trends, customer data, and competitor strategies.<br>• AI-Powered Chatbots: AI virtual assistants handle customer inquiries, wine selection, and order processing.<br>• AI-Based Crowdfunding: AI-powered platforms help wineries fundraise for social and community initiatives.<br>• AI evaluates online reviews and customer sentiment.<br>• Examples:<br>• *Vinous (AI-driven wine market analysis)*.<br>• *WineBot (AI-powered chatbot for wine recommendations)*.<br>• *BM Watson NLP (AI for sentiment analysis)*.<br>• *MonkeyLearn (AI for stance detection)*. |

| Collaboration with Local Businesses (e.g., partnerships with hotels, restaurants) | AI-Enhanced Business Networking | • AI-Driven Partnership Optimization: AI identifies and enhances strategic collaborations.<br>• AI-Powered Sentiment & Stance Analysis: AI monitors reviews and consumer trends, aligning strategies with market demand.<br>• AI Chatbots for Community Engagement: AI chatbots facilitate real-time communication and support between wineries and business partners.<br>• Examples:<br>• *ReviewPro (AI-driven sentiment monitoring).*<br>• *CommuneBot (AI chatbot for local collaborations).*<br>• *Google AI NLP (customer review sentiment analysis).*<br>• *Hugging Face Transformers (stance detection models).* |
|---|---|---|
| Social & Cultural Participation (e.g., winery events, cultural engagement) | AI-Driven Cultural Engagement | • AI-Powered Event Planning: AI assists in automating winery event logistics.<br>• AI for Digital Cultural Heritage: AI reconstructs historical wine traditions from historical data.<br>• AI-Driven Social Media Analysis: AI tracks trends and optimizes audience engagement.<br>• Sentiment & Stance Analysis for Cultural Engagement: AI evaluates public perception of winery events.<br>• Examples:<br>• *EventAI (AI-powered event planning and scheduling).*<br>• *HeritageAI (AI for preserving wine cultural heritage).*<br>• *SocialTrend (AI-driven social media analytics).*<br>• *Sprinklr AI (social media sentiment analysis).* |

*Source*: own elaboration based on: Alturayeif, Luqman & Ahmed, 2023; Alturayeif, Luqman & Ahmed, 2024; Saha, Lakshmanan & Ng, 2024; Mishra & Lourenço, 2024; Sprinkl, 2025; Yousaf, 2025; Hugging Face, 2025; AgriChain, 2025; Taranis, 2025; Google Cloud, 2025; Maar, 2025; Wine Industry Advisor, 2025; Shapes, 2025; Reviewpro Reputation, 2025; Medallia, 2025; WineBot, 2025; Ibm, 2025.

### 4.4 Wine tourism in Poland

Analysis of websites and social media portals reveals that the majority (ca. 80%) of registered Polish vineyards provide an option for tourist visits. As much as 83% of vineyards taking part in the survey declare openness to visitors and another 12% intend to include such an opportunity in the future, and only 5% of respondents indicate that they are not interested in providing this type of service.

Winemakers who offer the opportunity for tourists to visit a vineyard, i.e. 83% of respondents, were asked about the services they provide as part of their enotourism offer. The three most frequently provided services are wine tasting and purchase opportunity (58 indications), tour of the vineyard accompanied by the winemaker (57 indications) and guided tour of the winery and other buildings (e.g. wine ageing cellar) by the winemaker (44 indications). One in three wineries with a tourism function offers accommodation for tourists, workshops, training courses, lectures, speeches, and shows (20 indications each). The fewest wineries offer self-guided tours of the winery and other facilities (5 indications), spa/wellness services (3 indications) and packages with other service providers, for example a winery tour combined with a visit to another local tourist attraction (only 1 indication).

Respondents were also asked about the benefits of enotourism activities. This question was answered by 71 out of 75 respondents. The winemakers considered the most important benefits to be increased wine sales (58 indications), brand promotion (57 indications), additional sources of income, and strengthening the winemaker's social relations, including contacts, relationships, cooperation, and image (48 indications each). The least frequently reported benefit appeared to be financial security in the case of emergency events through diversification of business activities (only 1 indication). Two respondents found it difficult to indicate any benefits of enotourism activities.

The benefits of enotourism were complemented by an open-ended question on respondents' reasons for undertaking/planning an enotourism activity. This question was answered by 60 out of 75 respondents. The responses received can be categorized into the following content areas:

A1. Economic issues: wine sales (16 indications)
A2. Economic issues: additional income, e.g. from paid tastings (7 indications)
B. Promotional issues: vineyard, Polish winemaking (13 indications)
C1. Social issues: social relations (15 indications)
C2. Social issues: passion, hobby (7 indications)
C3. Social issues: feedback from customers, e.g. on wine quality (2 indications)
D. Development issues: diversification and expansion of activities (8 indications)
Z. Other (4 indications)

According to the survey, the most frequent reasons for undertaking/planning enotourism activities by respondents include economic issues, understood as the possibility of additional wine sales, social issues, understood as the possibility of contact with people, and promotional issues, understood as the possibility of promoting one's vineyard, wine and Polish winemaking in the broadest sense. These results, coded as A1 and B, correspond with the highest number of indications, with the results received in the question on the benefits of enotourism activities.

Table 4 shows the proposed scope of AI applications and possible methods linked to the winemakers' area of interest in sustainable tourism.

**Table 4.** Possible scope and methods of AI application in sustainable wine tourism.

| Winemakers' areas of interest | Scope of AI applications | AI Methods & Tools |
|---|---|---|
| Economic issues: wine sales<br><br>Economic issues: additional income, e.g. from paid tastings<br><br>Promotional issues<br><br>Social issues: feedback from customers, e.g. on wine quality | AI Chatbots as Virtual Sommeliers | • Natural Language Processing (NLP) – chatbots analyze customer queries and provide answers about wines and tastings.<br>• Machine Learning (ML) based on user preferences – AI systems recommend wines tailored to customer tastes based on purchase history. |
| Development issues: diversification and expansion of activities | Intelligent Mobile Applications for Visit Planning | • Predictive Analytics – analyzing historical booking data to forecast vineyard occupancy.<br>• Geolocation and traffic analysis – AI collects GPS data and adjusts tourist routes to current conditions. |
| Development issues: diversification and expansion of activities | AI in Tourist Behavior Analysis | • Computer Vision (image recognition) – analyzing camera footage to identify tourist movement and detect overcrowding.<br>• Deep Learning for visit pattern analysis – AI analyzes tourist behavior in real-time and adjusts offerings accordingly. |
| Promotional issues: vineyard, Polish winemaking | Translation and Content Adaptation | • Neural Machine Translation (NMT, e.g., Google Translate AI) – automatic content translation into multiple languages.<br>• NLP-based recommendation systems – adjusting educational content to the tourist's knowledge level. |
| Economic issues: wine sales | Virtual Wine Tasting Simulations | • Virtual Reality and Augmented Reality (VR/AR) – AI generates realistic tasting experiences in VR. |

| | |
|---|---|
| Economic issues: additional income, e.g. from paid tastings | • AI-driven Sensory Analysis – AI systems analyze users' sensory reactions to optimize virtual tastings. |
| Development issues: diversification and expansion of activities | |
| Promotional issues | |

*Source*: own elaboration based on: Lewis, Gemma K., et al., 2021; Roach, 2023; Aivin, 2025; Lighthouse, 2025; Wagner, 2025; Sommelier Business, 2025; Maar, 2025.

## 5. STUDY LIMITATIONS AND FUTURE RESEARCH

The research presented in this study focuses on winemakers. A complete picture of the wine industry's social, economic, and environmental impacts requires including the perspective of other local actors (residents and local authorities) as well as tourists. Additionally, economic analysis of statistical data might prove to be useful. The real impact of viticulture and wine production on the environment should be demonstrated through research based in natural and earth sciences, such as long-term ecological observations of biodiversity, or studies of changes in soil and water quality. Another significant issue is the relationship between Polish winemaking and climate change. Warming temporarily improves the conditions for viticulture, but the intensification of extreme weather events in the future may negatively affect this crop. Adaptation to climate change is therefore another determinant of winemaking sustainability and an important field of research.

Future research should focus on longitudinal studies to monitor the evolution of Polish enotourism, notably the transition from small-scale family operations to potentially larger, commercial ventures. Studies could also investigate the perspectives of tourists visiting Polish vineyards, exploring their expectations, satisfaction levels, willingness to engage in sustainable tourism practices, and AI tools. Future research on AI in sustainable viticulture should focus on developing agriculture techniques, AI-driven climate adaptation strategies, and intelligent resource management systems to enhance vineyard resilience, optimize yields, and minimize environmental impact. Similarly, research on sustainable wine production should explore waste reduction and organic production technologies to minimize environmental impact. Additionally, further analysis of the role of digital technologies, such as virtual vineyard tours or online wine tastings, in promoting sustainable tourism could provide insights into modernizing the sector.

## 6. DISCUSSION AND CONCLUSIONS

The questionnaire survey results show that Polish winemakers' activities fit in well with the crucial goals of sustainable development. Regarding the social pillar of sustainability, the Polish wine industry contributes to building social capital in localities where the vineyards are located. Social relations are important for winemakers. Most of them establish good relations with the people living near the winery and can show many examples of positive cooperation. Social relations were also an important benefit of a tourist offer. Winemakers' activities support the maintenance of local traditions (intangible heritage), among them culinary traditions (promotion of local meat and dairy products, local dishes). Winemakers willingly share the passion and knowledge they have gained during vineyard visits and commented that wine tastings are not limited to information about the wines and vines themselves. Winemakers also disseminate the knowledge about the region and its cultural heritage. Winemakers often become local animators of culture, organizing cultural events in wineries (mainly concerts, picnics, festivals). Tourism products created by winemakers can potentially increase the popularity of the location of the vineyard. Polish wine brands often incorporate geographical designations – names of villages or regions where they are produced and their wine labels frequently feature distinctive symbols of the locality or region. This serves as an additional form of promotion, contributing to greater recognition of the areas where vineyards are located. Many vineyards are situated in regions not traditionally considered particularly attractive for tourism. Therefore, the development of enotourism offerings can act as a magnet, attracting visitors to these areas and contributing to the deglomeration of tourist traffic.

Regarding the economic pillar of sustainability, the Polish wine industry and wine tourism create additional employment opportunities in rural areas and contribute to developing non-agricultural functions of rural areas. Most winemakers indicated that they contribute to improving local development and well-being. This mostly includes providing employment for the local people, purchasing from local stores or entrepreneurs, buying local products, and using local services. Wine tourism supports local suppliers of products and services, thus diversifying sources of income and providing opportunities to earn money beyond the main season of agricultural activities.

In the environmental pillar of sustainability, winemakers are aware of the significance of biodiversity and current trends in wine production – ecological and organic winemaking – and try to reduce their negative environmental impact. Environmental awareness in the broadest sense includes landscape protection, so winemakers are usually aware of the value of rural cultural landscapes. They aim to preserve traditional architecture and other cultural heritage elements, so, for example, when building new infrastructure, they try to follow local style and tradition. The question remains whether the sustainability of Polish winemaking, viticulture and wine tourism is just a matter of scale, and thus whether it will be lost with the shift to mass production.

Polish winemakers are increasingly aware of the impact of climate change, which both benefits and challenges viticulture. While warming temperatures have facilitated the re-emergence of vineyards, the growing frequency of extreme weather events poses risks to crop yields and long-term sustainability. Adaptation strategies, such as cultivating disease-resistant grape varieties, improving water management systems, and investing in renewable energy, are critical for mitigating these challenges. Further research should assess how these strategies can be systematically implemented across the industry.

The first question is whether such a shift toward mass production of wine will occur. Another problem is whether the principles of sustainable development can carry out such a transformation. Whether the next generations will take over wine activity, or in what form and approach is uncertain. Mass tourism is becoming increasingly troublesome for local communities and is facing protests, so developing enotourism into a mass trend is not a foregone conclusion. Since we are dealing with a new phenomenon that is developing under conditions of significant climatic (increasing likelihood of extreme weather events), social (likely demographic crisis) and economic (unstable or rising energy prices) uncertainty, it must be assumed that various development scenarios of Polish viticulture and related wine tourism are possible.

Artificial Intelligence (AI) has the potential to support the sustainable development of Polish winemaking by providing tools for optimizing production processes, reducing resource consumption, and increasing the efficiency of enotourism activities. Predictive systems based on machine learning can assist in managing irrigation and crop protection, minimizing the use of chemical agents, and reducing environmental impact. AI can also help winemakers monitor biodiversity and implement eco-friendly cultivation methods.

In the social aspect, AI tools can support the promotion of local traditions and products through personalized recommendations for tourists and automated interaction systems, such as chatbots acting as virtual sommeliers. In the economic context, intelligent data analysis algorithms can help optimize supply chains, reduce losses, and tailor offerings to changing consumer preferences.

Regarding the natural environment, it is crucial to implement solutions that minimize the negative impact of winemaking on ecosystems. AI can support soil and groundwater quality monitoring, helping to limit erosion and chemical pollution. Intelligent image analysis systems can be used to identify plant diseases at an early stage, allowing for a reduction in pesticide and fungicide use. Additionally, predictive technologies can aid vineyard adaptation to climate change by selecting more resistant grape varieties and optimizing microclimate management within vineyards.

In conclusion, implementing AI solutions in Polish winemaking can contribute to the industry's further development in line with sustainability principles. However, it will be essential to tailor technology to the specifics of local vineyards and ensure adequate support for winemakers so that innovative solutions are accessible and effectively implemented. Future research should continue to examine the long-term impact of AI on the sustainability of winemaking, particularly in climate change and its consequences for Polish vineyards.

Most winemakers in Poland are not currently using advanced technologies, including AI to support the management of viticulture, wine production and enotourism. Perhaps the demand for such solutions will increase when the size of vineyards, the scale of production and the number of tourists grow significantly.

**Funding:** This research was co-financed by the Minister of Science under the "Regional Excellence Initiative".